\documentclass[letterpaper]{article} 
\usepackage{style}  
\usepackage{times}  
\usepackage{helvet}  
\usepackage{courier}  
\usepackage[hyphens]{url}  
\usepackage{graphicx} 
\usepackage{natbib}  
\usepackage{caption} 
\usepackage{algorithm}
\usepackage{algorithmic}
\usepackage{booktabs}
\usepackage{bbding}
\usepackage{amssymb}
\usepackage{url}

\usepackage{xcolor}

\usepackage{tcolorbox, soul}

\usepackage{newfloat}
\usepackage{listings}
\DeclareCaptionStyle{ruled}{labelfont=normalfont,labelsep=colon,strut=off}
\floatstyle{ruled}
\newfloat{listing}{tb}{lst}{}
\floatname{listing}{Listing}

\title{Can Large Models Fool the Eye? A New Turing Test for Biological Animation}
\author {
    Zijian Chen\textsuperscript{\rm 1,\rm 2},
    Lirong Deng\textsuperscript{\rm 3},
    Zhengyu Chen\textsuperscript{\rm 1},  
    Kaiwei Zhang\textsuperscript{\rm 2},
    Qi Jia\textsuperscript{\rm 2},
    Yuan Tian\textsuperscript{\rm 2}, \\
    Yucheng Zhu\textsuperscript{\rm 1},
    Guangtao Zhai\textsuperscript{\rm 1,\rm 2,\rm *}
}
\affiliations {
    \textsuperscript{\rm 1}Shanghai Jiao Tong University, Shanghai 200240, China\\
    \textsuperscript{\rm 2}Shanghai AI Laboratory, Shanghai 200232, China\\
    \textsuperscript{\rm 3}Macao Polytechnic University, Macao 999078, China\\
    \textsuperscript{\rm *}Corresponding author
}

\usepackage{bibentry}

\begin{document}

\maketitle

\begin{abstract}
Evaluating the abilities of large models and manifesting their gaps are challenging. Current benchmarks adopt either ground-truth-based score-form evaluation on static datasets or indistinct textual chatbot-style human preferences collection, which may not provide users with immediate, intuitive, and perceptible feedback on performance differences. 
In this paper, we introduce BioMotion Arena, a novel framework for evaluating large language models (LLMs) and multimodal large language models (MLLMs) via visual animation.
Our methodology draws inspiration from the inherent visual perception of motion patterns characteristic of living organisms that utilizes point-light source imaging to amplify the performance discrepancies between models.
Specifically, we employ a pairwise comparison evaluation and collect more than 45k votes for 53 mainstream LLMs and MLLMs on 90 biological motion variants. 
Data analyses show that the crowd-sourced human votes are in good agreement with those of expert raters, demonstrating the superiority of our BioMotion Arena in offering discriminative feedback.
We also find that over 90\% of evaluated models, including the cutting-edge open-source InternVL3 and proprietary Claude-4 series, fail to produce fundamental humanoid point-light groups, much less smooth and biologically plausible motions. 
This enables BioMotion Arena to serve as a challenging benchmark for performance visualization and a flexible evaluation framework without restrictions on ground-truth.
\end{abstract}

\begin{links}
    \link{Project}{https://zijianchen98.github.io/BioMotion-Arena/}
\end{links}

\section{Introduction}
Large language models (LLMs) and multimodal large language models (MLLMs) have shown groundbreaking performance across various applications, including routine visual question answering \cite{bai2025qwen25,zhu2025internvl3}, native audio outputs \cite{gemini2.5}, code \cite{guo2024deepseek}, mathematics \cite{gulati2024putnam}, and multidisciplinary tasks \cite{yue2024mmmu,chen2024obi}, necessitating enhanced evaluation approaches to keep up with their rapid advancements.
Currently, most benchmarks often fail to convey subtle performance differences and align with user preferences in real-world scenarios.

\begin{figure}[t]
\begin{center}
\includegraphics[width=1\linewidth]{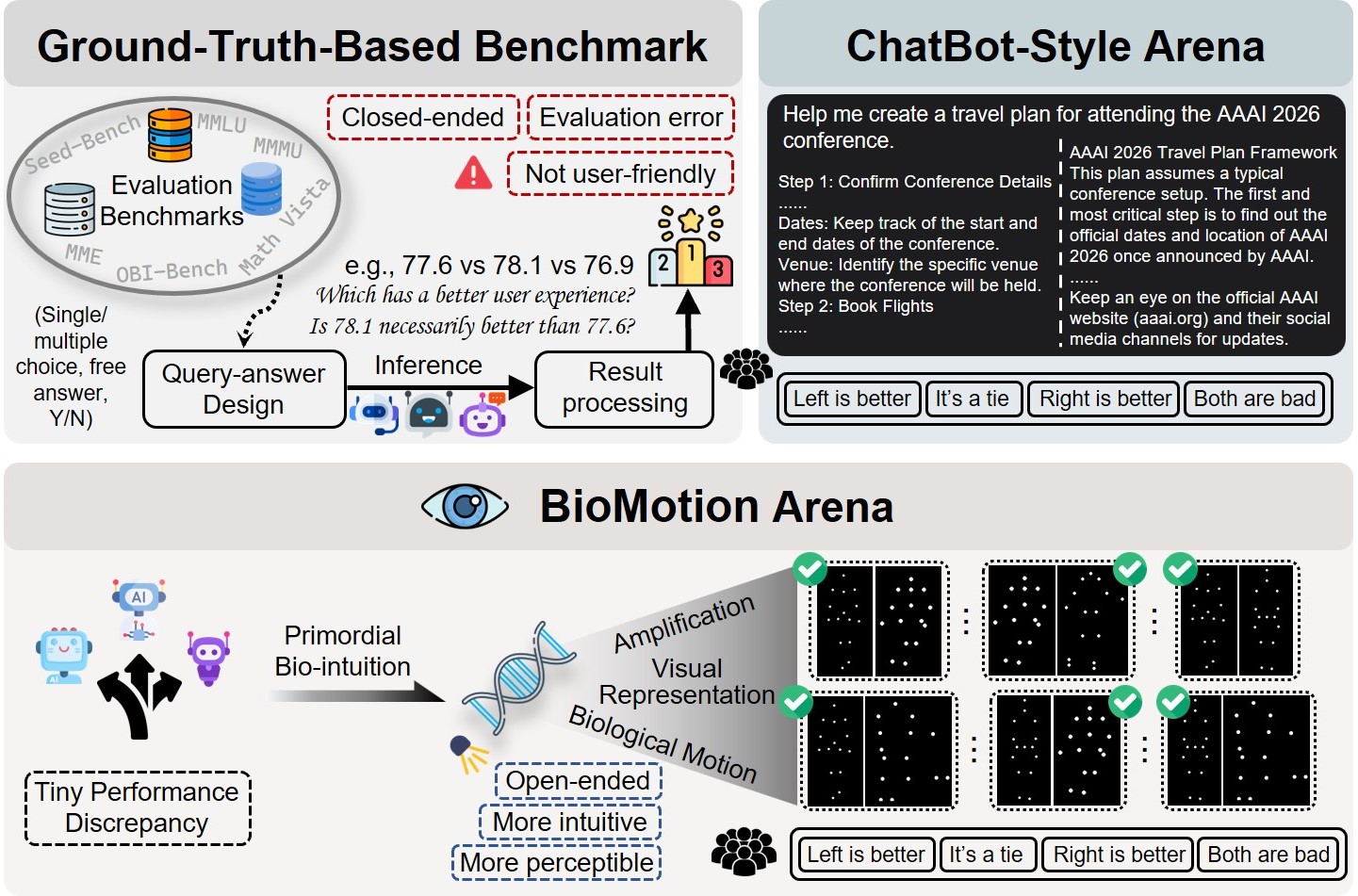}
\caption{Comparison of existing large model benchmarks and our {\bf BioMotion Arena}. We categorize around the evaluation presentation formats: whether the evaluation relies on ground truth-guided numerical comparison or text-oriented human performances. BioMotion Arena is the first to incorporate the perception of fundamental biological motion into visual preference-based human evaluation.}
\label{intro}
\end{center}
\end{figure}

The mainstream benchmarks can be classified into two categories: the ground-truth-based closed-ended benchmark and the human preference-oriented open-ended benchmark, as shown in Fig. \ref{intro}. The former relies on meticulous dataset construction and task design but is susceptible to data contamination and incurs high maintenance costs.
The latter, especially those text response-based human preference evaluations, such as Chatbot Arena \cite{chiang2024chatbot}, Arena-Hard \cite{licrowdsourced}, and GameArena \cite{hugamearena}, may be affected by subjective bias (e.g., age, gender, and educational background) given the noisy nature of human preferences \cite{lehmann2020interaction}.
Moreover, the closeness of scores and output formats can hardly convey performance disparities across models to users intuitively. Consequently, it is imperative to build an open, intuitive, unbiased evaluation framework that can well present the performance.

\begin{figure*}[t]
\begin{center}
\includegraphics[width=1\linewidth]{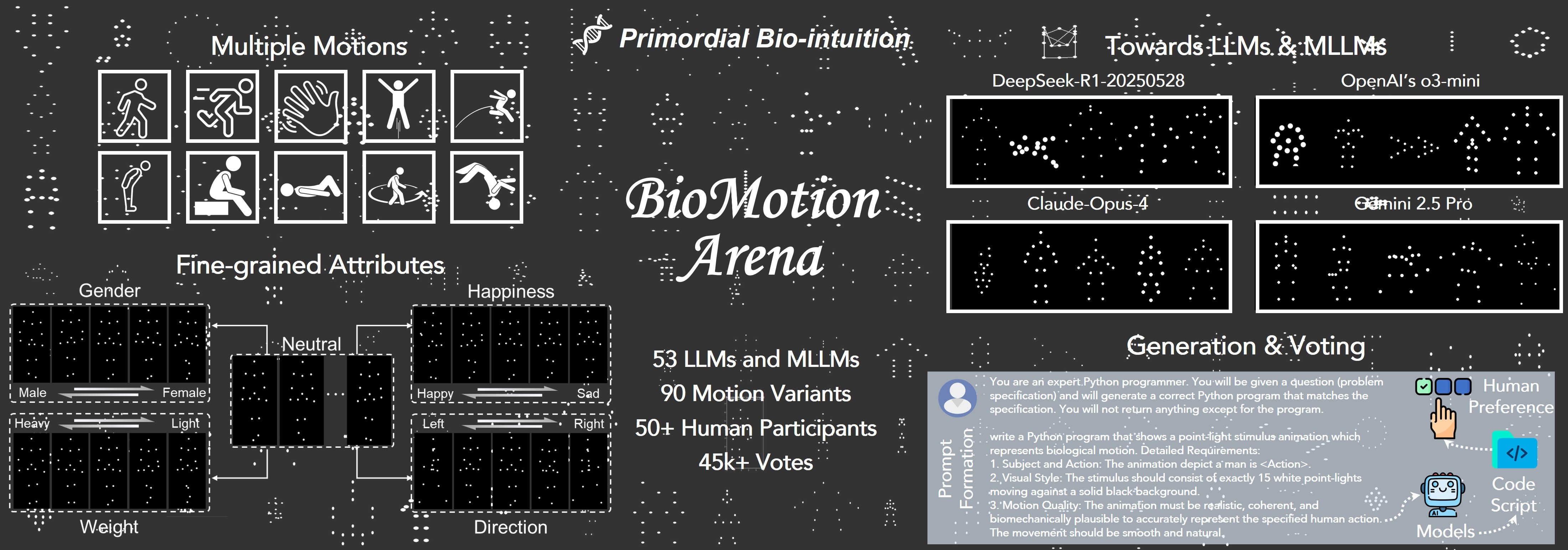}
\caption{Overview of BioMotion Arena. It is an open platform for evaluating large models based on visual content-guided human preferences, featuring multiple motion and attribute types for biologically animating. The lower left part visualizes some {\it golden} reference point-light walkers across different human gait patterns. 15 markers are located at the major joints of the body (shoulders, elbows, wrists, hips, knees, ankles), the sternum, the center of the pelvis, and the center of the head.}
\label{framework}
\end{center}
\end{figure*}

To this end, we propose BioMotion Arena, a visual benchmarking approach for large models that features easily perceptible and ground-truth-free comparisons in a lightweight setting. 
BioMotion Arena draws inspiration from a stunning example of the resourcefulness of human vision, namely the remarkable capacity to perceive `biological motion', which has a far higher degree of complexity than the patterns of simple mechanical motions and was first shown in earlier cinematic works \cite{neri1998seeing}. 
A point-light display, illuminating only the major joints of a walking person, is sufficient to generate a vivid and coherent percept of human animation. This perception, however, disintegrates into a jumble of meaningless dots when the walker remains stationary. It is worth noting that such a phenomenon happens in young infants \cite{fox1982perception} and can be interpreted easily without any knowledge foundation.

Armed with this, we first focus on ten typical human motions and introduce fine-grained control over gender, weight, mood, and direction. Concretely, in BioMotion Arena, a user can prompt to generate a human animation that satisfies the above combination of action requirements and get visual answers from two anonymous large models.
Afterwards, the user is instructed to vote for the model that delivers the preferred animation.
Such a lightweight setup enables efficient collection of diverse prompts on varying motions and simplifies the process of comparing different models using visual preference data, providing an intuitive presentation to evaluate a large model's grasp of the underlying mechanisms of biological motion.

We have conducted extensive experiments with 53 LLMs and MLLMs, where over 50 human annotators participated, with more than 45k votes collected. Based on this, we construct a leaderboard by translating pairwise comparisons into ranking through Elo score \cite{elo1967proposed}, identifying the state-of-the-art models, such as DeepSeek-R1-20250528, Gemini 2.5 Pro, and OpenAI's o3, in animating biological motions.  
The winning fraction heatmaps reveal that the reasoning or thinking mode in some newly released models can improve the realism of generated motions.
We also performed several case studies for qualitative analysis, demonstrating that existing models are still far from generating vivid, smooth, and realistic biological motions.
We are committed to making BioMotion Arena an open-access online evaluation platform and releasing the preference data periodically.

\begin{figure*}[t]
\begin{center}

\includegraphics[width=1\linewidth]{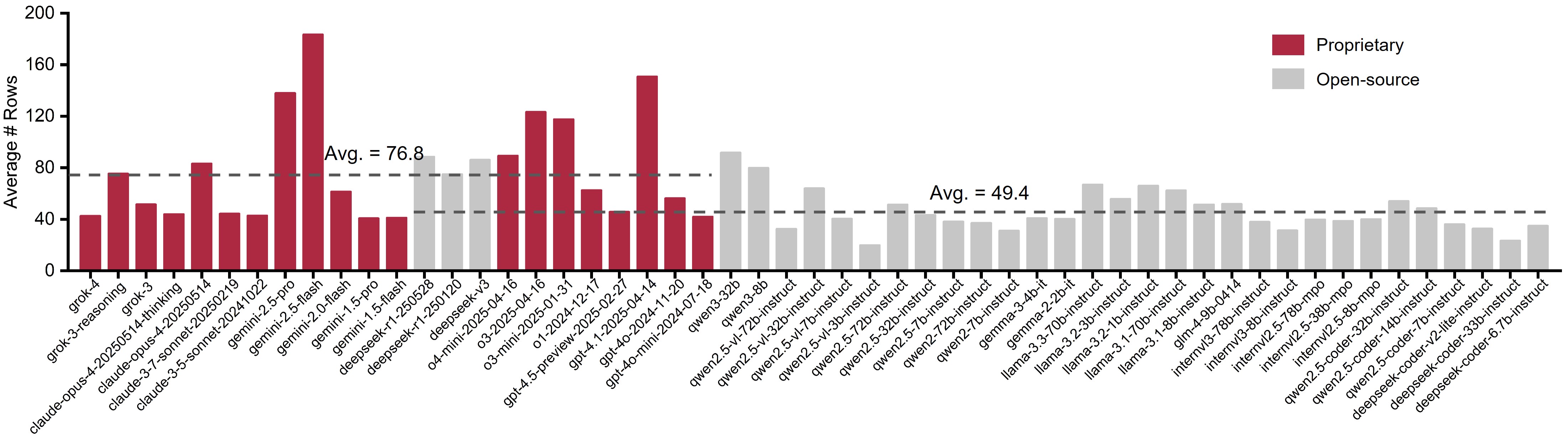}
\caption{Statistics of the average lines of code for biological motion representation generated by different models.}
\label{code-length}
\end{center}
\end{figure*}

Our contributions can be summarized as follows:
\begin{itemize}
    \item We propose the first biological motion-based visual preference evaluation framework for large models.
    \item We conduct an in-depth analysis of the collected data, including the generated animation, win-rates across model categories, inter-motion correlations, and vote quality.
    \item We will continuously release human preference data with valid pairwise votes collected from BioMotion Arena. 
\end{itemize}

\section{Related Work}

\noindent
{\bf LLM and MLLM Benchmarks.}
We briefly review the current LLM and MLLM benchmarks following the taxonomy presented in Fig. \ref{intro}.
Prevailing benchmarks for LLMs and MLLMs, spanning diverse evaluation dimensions such as visual perception \cite{li2024seed,liu2024mmbench}, text comprehension \cite{wang2024mmlu}, multidisciplinary knowledge \cite{yue2024mmmu,chen2024obi,aibench}, reasoning \cite{valmeekam2023planbench,gulati2024putnam}, and safety \cite{li2024salad}, are predominantly static and closed-ended, making them highly susceptible to data contamination. More importantly, most of them rely on multiple-choice and binary question-answer format designs, which are vulnerable to model hallucinations and output biases \cite{chen2025just}.
Later, live benchmarks with data bootstrapping strategies \cite{zhang2025puzzlebench,yang2024dynamic} or dynamically updated questions \cite{whitelivebench,jainlivecodebench} have been proposed to address such problems.
However, these studies are evaluated based on quantitative scores that are not only challenging to convey discrepancies but also poor indicators of real-world user experience.
The recent trend involves utilizing live human interaction for benchmarking the performance of large models, with notable instances including Chatbot Arena \cite{chiang2024chatbot}, Arena-Hard \cite{licrowdsourced}, GameArena \cite{hugamearena}, and WV-Arena \cite{luwildvision}.
A significant limitation is that the resulting subjective preferences from these platforms are primarily text-centric, lacking the consideration of task-oriented scenarios and other modalities (e.g., image or video).
In this paper, we propose to leverage the foundational capability of biological visual perception to achieve a more intuitive differentiation of model performance.

\noindent
{\bf Human Preference Dataset.} To bridge the gap between model outputs and user expectations, generative AI development increasingly relies on human preference annotations, which guide the training process, steering models towards desired behaviors, and have become integral to advancements in image synthesis \cite{kirstain2023pick,wu2023human}, web search optimization \cite{nakano2021webgpt}, and enforcing safety constraints \cite{bai2022training,ji2023beavertails}.
In the field of language model development, several human preference datasets, such as OpenAssistant \cite{kopf2023openassistant}, LMSYS-Chat-1M \cite{zhenglmsys}, and AMP \cite{zhang2024automated}, have been proposed to facilitate MLLMs' evolvement towards learning superior responses while avoiding inferior ones.
However, most of these works focus solely on conversations and lack human preference data for other modalities, hindering their direct application in ranking other output formats.

\begin{figure*}[t]
\begin{center}
\includegraphics[width=1\linewidth]{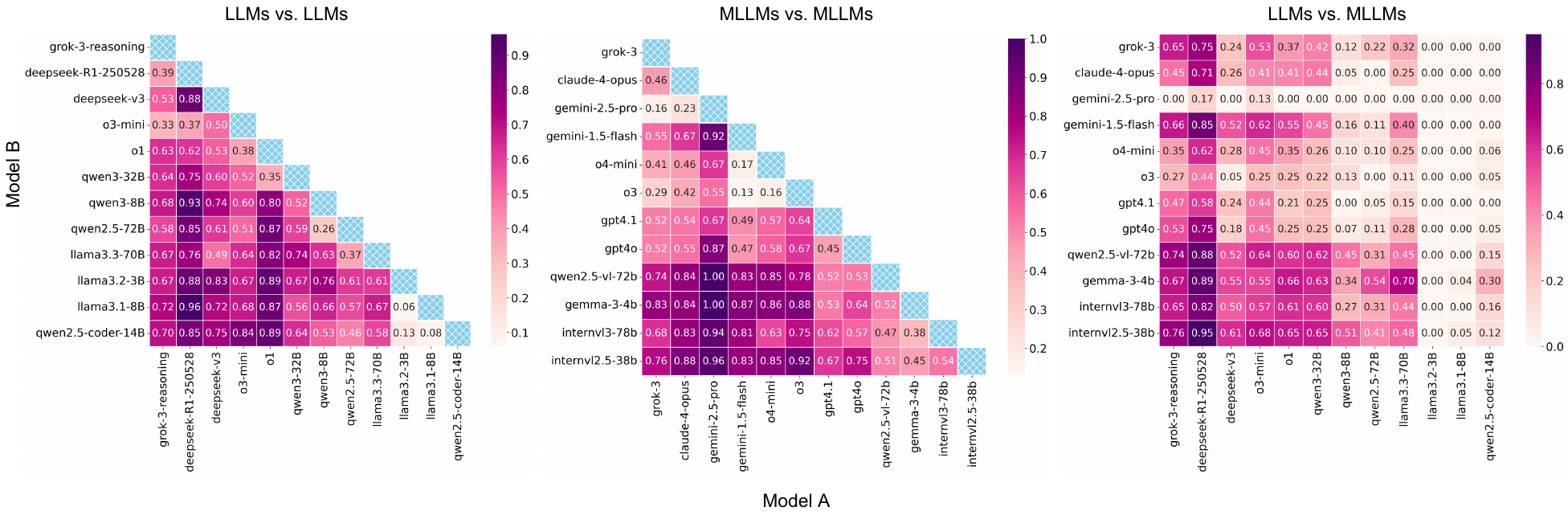}
\caption{Win-rate (Model A beats Model B) between a subset of models in BioMotion Arena. }
\label{win-rate}
\end{center}
\end{figure*}

\section{The BioMotion Arena}

\subsection{Overview}
BioMotion Arena aims to evaluate the understanding of biological motion patterns of large models and is constructed in response to two problems. First, the current benchmarks predominantly present model performance using single numerical scores, failing to effectively convey the nuanced ability differences between models to users and developers. Second, the existing subjective preference datasets, particularly those collected from chatbot-style systems, primarily rely on votes on textual content. 
Such a scheme is highly susceptible to biases stemming from annotators' diverse backgrounds (e.g., age, education, and cognitive level), thereby increasing inter-group variance and introducing noise into the preference data.

Specifically, we gain inspiration from the fundamental biological perception mechanism that human vision has the ability to see `biological motion’. For example, illumination of only the joints of a walking person is enough to convey a vivid, compelling impression of human animation, which is sufficient to discriminate the sex and other details of the walker, and can be easily interpreted by young infants \cite{neri1998seeing}.
Previous studies \cite{johansson1973visual,thornton2006biological} have demonstrated that the human visual apparatus is very sensitive to small deviations from such standard patterns, as we can immediately see whether a person is walking, running, dancing, or moving forward.

In this paper, we simulate different biological motions by generating point-light stimuli with code prompts, in which each dot represents specific joints of a human performing an action. Grounded in fundamental biological perception, the subjective feedback of the resulting human gait animations represents the performance of the model evaluated, offering a more intuitive mode of judgment than the textual format. 
Fig. \ref{framework} provides an overview of BioMotion Arena.

\subsection{Biological Motion Generation}

\noindent
{\bf Motion and Posture Space.} 
Following the definition of point-light stimuli that the body posture is encoded in the location of the point-lights \cite{theusner2014action}, we mainly focus on ten typical human actions with pronounced limb movements, including walking, running, waving a hand, jumping up, jumping forward, bowing, lying down, sitting down, turning around, and forward rolling. 
Moreover, since biological motion contains information about several different emotions, intentions, personality traits, and biological attributes of the agent \cite{troje2002decomposing, troje2008retrieving}, which can be retrieved by a person at a glance, we further incorporate finer-grained dimensions, including gender, weight, happiness, and direction. This poses stricter scenarios for large models to understand and generate.

To obtain the {\it golden} reference motion animation, we utilize the joint kinematics function formulated in \cite{troje2002little}, which is fitted based on the real walking data from 50 male and 50 female walkers.
Their posture visualization as a point-light walker is shown in Fig. \ref{framework} (lower-left part). We can observe that the sexual dimorphism in human body proportions is evident, with males demonstrating significantly greater biacromial breadth and a wider gait stance compared to females. In terms of body weight, people with a higher weight tend to exhibit greater distances between corresponding points on the upper and lower limbs, which is particularly pronounced in the waist region.
Besides, people with a happier mood tend to exhibit a greater amplitude in their arm swing during locomotion, whereas individuals experiencing sadness display a pronounced slumping or dropping of the shoulders.
Beyond action categories, we also treat the number of point-lights as a variable, given its critical impact on the perceived coherence of the biological motion \cite{peng2021exploring}. See more visualizations in the Appendix.

\noindent
{\bf Prompt Design.} The BioMotion Arena is fundamentally a code generation scenario that follows the standard setup for generating code from natural language \cite{jainlivecodebench}. 
Specifically, the template begins with a system prompt of the role assigned to the model, followed by the inclusion of point-by-point detailed requirement texts with interleaved reference images. Then the model is tasked with generating an appropriate solution for visually vivid biological motions.
For LLMs and MLLMs, we adopt the text-only and image-text prompting, respectively. One frame from the golden reference motion of forward walking is used as the input image.
The lower right part of Fig. \ref{framework} provides an example scenario for evaluating LLMs.

\noindent
{\bf Benchmark Candidates.} Our BioMotion Arena currently includes 53 large models (both LLMs and MLLMs) in total, with a mix of cutting-edge proprietary models, open-source models, and code-specific models. In particular, for proprietary models, we include OpenAI models such as o4-mini, GPT-4.5-Preview, and GPT-4o \cite{GPT-4o,GPT4.5,o4}, Google models such as Gemini-2.5-Pro and Gemini-2.0-Flash \cite{gemini2.5,gemini2}, Anthropic models such as Claude-Opus-4 and Claude-3-7-Sonnet \cite{claude37,claude4}, and xAI models such as Grok-4 and Grok-3 \cite{grok}.
For open-source models, we comprise models including Qwen3 \cite{yang2025qwen3}, Qwen2.5-VL \cite{bai2025qwen25}, Gemma3 \cite{team2025gemma}, Llama3.x \cite{grattafiori2024llama}, and Internvl3 \cite{zhu2025internvl3} series.
Additionally, we select two code-specific LLM series, i.e., Qwen2.5-Coder \cite{hui2024qwen2} and Deepseek-Coder \cite{guo2024deepseek}, for evaluation. The \texttt{max\_new\_tokens} is set to 32768 for all models to avoid performance degradation caused by insufficient responses.
Fig. \ref{code-length} plots the average lines of code for biological motion representation generated by different models.
Furthermore, apart from the proprietary LMMs that are deployed via API, all other models are performed using a maximum of 8 Nvidia A800 80GB GPUs.
See the Appendix for a full list of models and citations.

\begin{figure}[t]
\begin{center}
\includegraphics[width=1\linewidth]{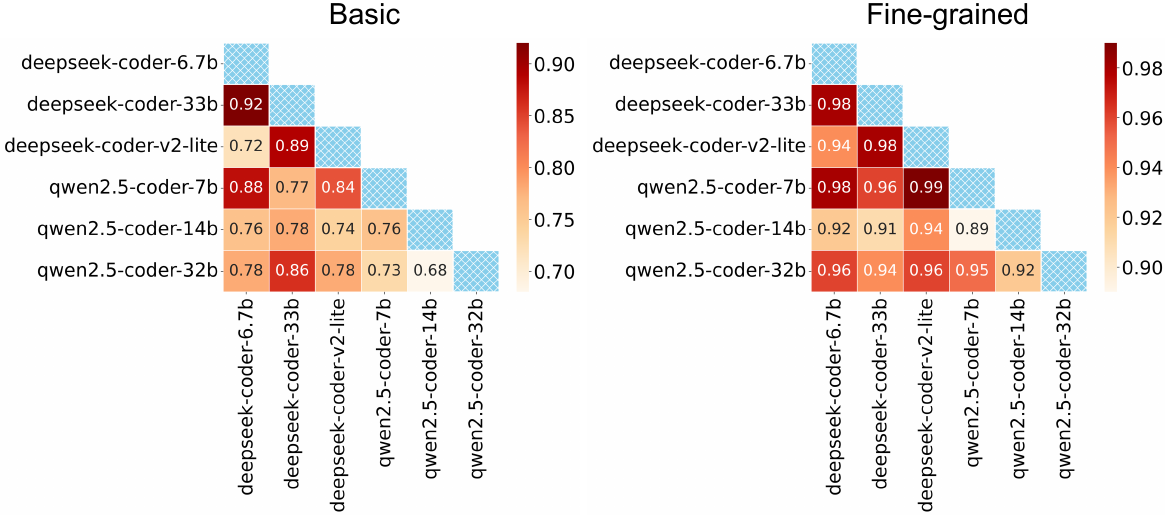}
\caption{The rate of `Both-are-bad' within the code-specific comparisons.}
\label{bad-rate}
\end{center}
\end{figure}

\subsection{Human Preference Data Collection}
\noindent
{\bf User Interface.} 
BioMotion Arena aims to collect dynamic visual content-based feedback from users for model evaluation. 
Different from the common quality assessment for AI-generated content \cite{chen2024gaia,chen2025study} that employs unified grading criteria or continuous scales, we adopt a pairwise comparison mechanism, where users are instructed to compare the output animation and vote for the better one.
A screenshot of the interface is shown in the Appendix. 
To avoid subjective bias towards the models themselves, in each round, we only provide the target type of the generated action while concealing the specific model from which it originated.
Two code scripts are randomly sampled and executed sequentially, and then users compare the generated biological motions side-by-side and vote for the preferred one. We also add two neutral buttons, i.e., ``it's a tie" and ``both are bad", for ambiguous pairs and failed scenarios.
In the future, we plan to make this platform online to better collect human votes longitudinally.

\noindent
{\bf Participants.} We recruited 53 participants from universities and society through hybrid online–offline channels to collect sufficient human votes.
All of them have normal or corrected eyesight with no color blindness. Before annotations, we obtained a signed and informed agreement from each person to address the ethical concerns for the usage of subjective preference data.

\noindent
{\bf Translating Pairwise Comparisons into Ranking.}
To construct a dynamic, predictive, and sustainably updated leaderboard based on BioMotion Arena, we use the Elo score \cite{elo1967proposed}. Its mathematical foundation rests on two key formulas. First, the expected score ($E_{A}$) of model A against model B is calculated based on their current ratings, $R_A$ and $R_B$, using a logistic function:
\begin{equation}
    E_{A}=\frac{1}{1+10^{\left( R_{B}-R_{A} \right) /400}}.
\end{equation}
The expected score for model B can be calculated by $E_B = 1-E_A$. After a round of comparison, the ratings are updated based on the actual outcome ($S_A$) compared to the expected score ($E_A$):
\begin{equation}
    R_{A}^{\prime}=R_{A}+K\cdot \left( S_{A}-E_{A} \right) ,
\end{equation}
where $K$ is a constant that determines the magnitude of the score adjustment. 
As a result, a win battle over a lower-rated opponent yields fewer points for the higher-rated model than what is lost in a defeat.
In this work, all models are initialized with a rating of $1500$, and $K$ is set to $32$.

\begin{figure}[t]
\begin{center}
\includegraphics[width=0.95\linewidth]{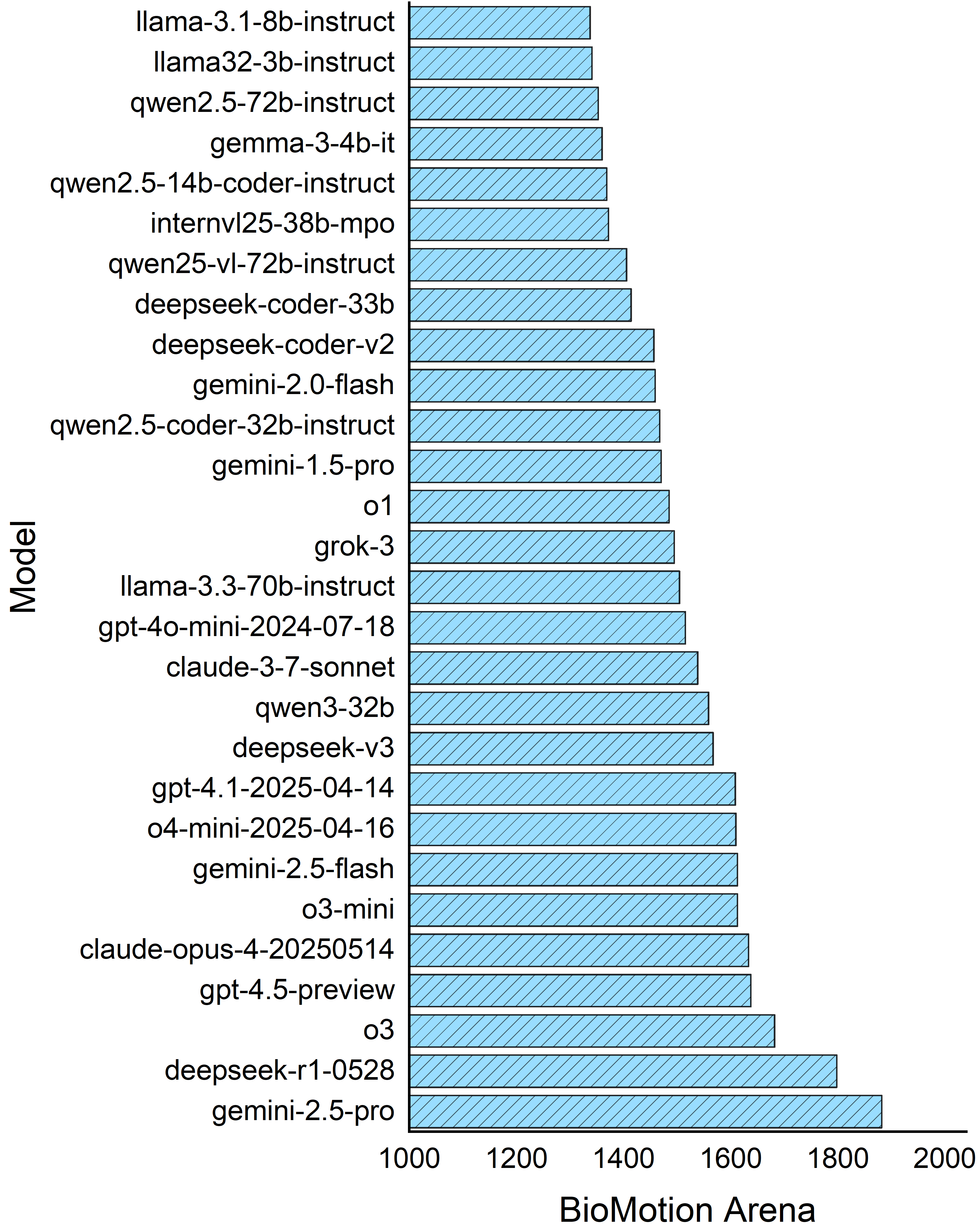}
\caption{Elo scores of a subset of models in the BioMotion Arena, showing a significant gap between open-source and proprietary models.}
\label{elo-score}
\end{center}
\end{figure}

\section{Data Analysis}

\subsection{Main Results}

\noindent
{\bf Pairwise Comparison.} 
In Fig. \ref{win-rate}, we visualize the heatmap of win fractions of twelve representative LLMs and MLLMs out of the 53 models supported in the BioMotion Arena. For battles between LLMs, deepseek-R1-0528 consistently outperforms the others (except for OpenAI's o3-mini) by a large margin, winning 62\% of its battles against the second-best model, OpenAI's o1.
Among the other open-source LLMs, DeepSeek-V3 and Qwen3-32B relatively lead, though there still remains a significant gap between them and the proprietary models.
For battles between MLLMs, where reference images are provided during motion generation, Gemini 2.5 Pro establishes a commanding lead over the entire field of competitors. Notably, it obtained a perfect win rate in its matchups against both Qwen2.5-VL-72B and Gemma-3-4B.
Claude 4 Opus and o3 follow closely behind Gemini 2.5 Pro, winning 23\% and 45\% of their respective battles, while other models demonstrate lower winning rates. 
We also conducted cross-modality comparisons between LLMs and MLLMs, as shown in the right part of Fig. \ref{win-rate}.
It can be seen that deepseek-R1-0528 exceeds most MLLMs while achieving only a 17\% and 44\% win rate of its matches against Gemini 2.5 Pro and o3, respectively.
In general, LLMs exhibit slightly inferior generative performance compared to MLLMs, primarily due to their lack of visual grounding. Nonetheless, their counterparts equipped with reasoning capabilities, such as grok-3-reasoning, o3-mini, and o1, continue to demonstrate impressive performance, showing their intrinsic understanding of the biological movement patterns.

\noindent
{\bf Poor Understanding of Biological Motions.} 
As shown in Fig. \ref{bad-rate}, the average occurrence rate of `Both-are-bad' in basic motions and fine-grained variants is up to 79.3\% and 94.8\% in the code-specific comparisons, respectively, indicating the severe inferiority of the existing code-specific LLMs in understanding the patterns of biological motions.
In Fig. \ref{elo-score}, we provide the leaderboard of BioMotion Arena for 28 models in terms of the Elo score. It can be observed that the BioMotion Arena effectively reveals a significant performance gap between proprietary and most open models. Besides, models featuring multi-step reasoning capabilities perform better than traditional LLMs, demonstrating the potential of such a strategy in enhancing the understanding ability of models. Similar conclusion can also be drawn from the length of the generated code, as shown in Fig. \ref{code-length}. 

\begin{figure}[t]
\begin{center}
\includegraphics[width=1\linewidth]{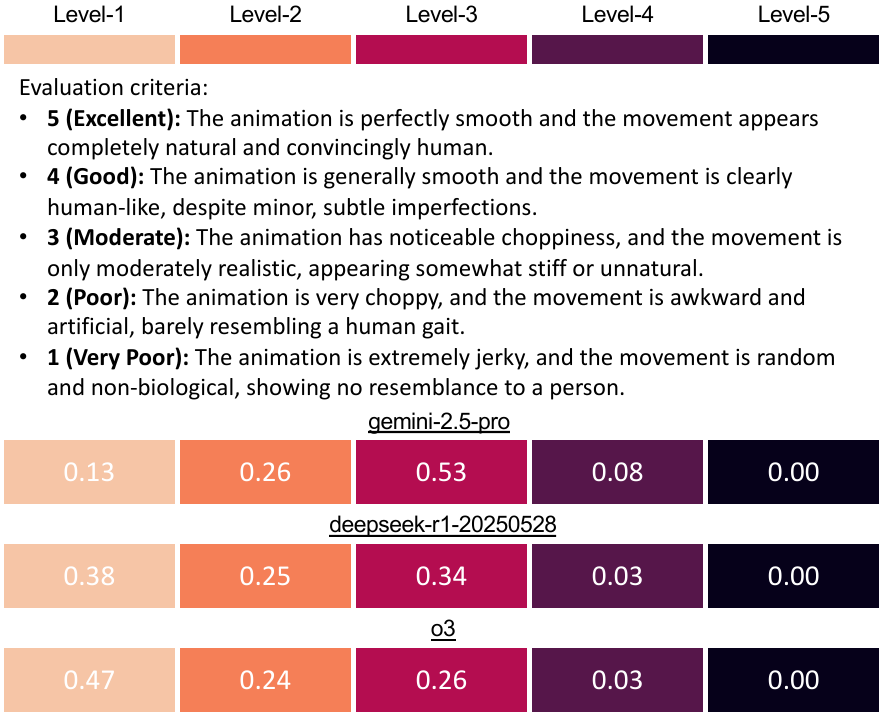}
\caption{The absolute rating distribution of the {\it top}-3 models in BioMotion Arena.}
\label{absolute-score}
\end{center}
\end{figure}

In addition to comparing the relative performance, we further conduct subjective experiments to investigate the absolute rating distribution. As shown in Fig. \ref{absolute-score}, we elaborate on a five-point Likert rating scale, and 10 subjects were invited to participate. We can see that even the {\it top}-3 models receive scores of only three or less, suggesting a significant gap remains in generating highly realistic biological motion, while also validating the challenging nature of our BioMotion Arena.

\noindent
{\bf Can BioMotion Arena Distinguish Models?}
We also study whether biological motion prompts in the BioMotion Arena can differentiate models. Designing challenging prompts has become increasingly difficult due to the rapid advancements of large models with potential data contamination. 
A case in point is that models with large parameters, such as Qwen2.5-72B and Llama3.3-70B-Instruct, are capable of solving most scenarios in daily life as good as GPT-4o, but not in domains such as math or coding.  
However, this phenomenon could, in turn, be upended by a later model like Qwen3, highlighting the importance of achieving consistent results in a standardized task.
Here, we sample the battles between Claude 3.7 Sonnet versus o4-mini-20250416 and GPT-4o-20241120 by segmenting them into ten actions and compiling statistics on their respective win rates and collapse rates, as summarized in Tab. \ref{claude37-win}.
We notice that Claude 3.7 Sonnet and its opponents exhibit significantly lower collapse rates in actions with low-amplitude movements, such as walking, waving a hand, and lying down, while failing more when encountering a greater range of motions.
Similarly, o4-mini achieves a higher win-rate in actions with slight movements ($\approx$ 95\%), which drops to about 30\% in vigorous motions.
In general, these results demonstrate the effectiveness of BioMotion Arena in distinguishing models' strengths.

\begin{figure*}[t]
\begin{center}
\includegraphics[width=0.95\linewidth]{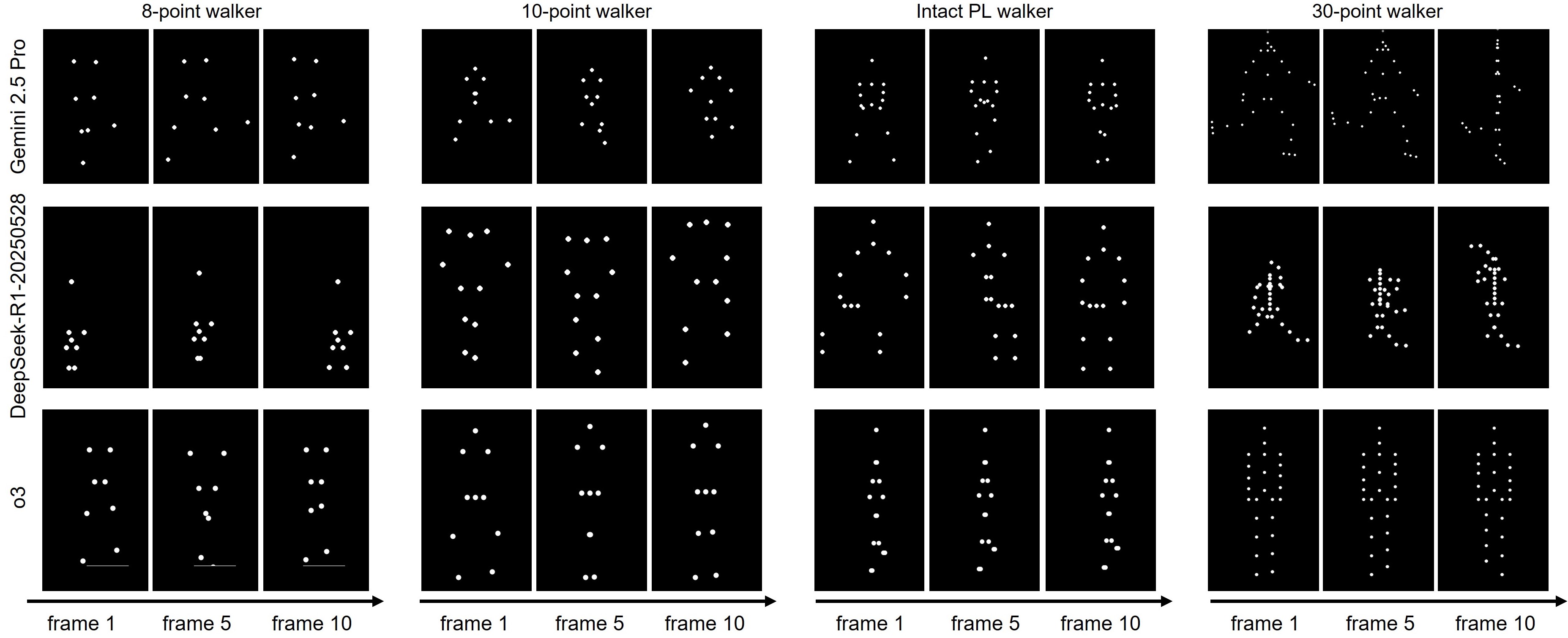}
\caption{Sample frames from an eight-point walker, a ten-point walker, an intact point-light walker, and a thirty-point walker. The intact point-light stimulus consists of 15 dots attached to the joints of a moving human figure.}
\label{few_points}
\end{center}
\end{figure*}

\begin{table}[t]
    \centering
    \renewcommand\arraystretch{1}
    \caption{Claude 3.7 Sonnet's win-rate against o4-mini-2025-04-16 and GPT-4o-2024-1120 on different motions. The values in parentheses denote the proportion of `Both-are-bad'.}
   \begin{tabular}{lcc}
    \toprule                 
        Motion&o4-mini&GPT-4o-2024-1120\\
        \midrule
        Walking&0.9\% (0.2\%)&25.4\% (24.9\%)\\
        Running&48.7\% (12.3\%)&62.6\% (25.3\%)\\
        Waving a hand&24.6\% (0.2\%)&50.7\% (12.5\%)\\
        Jumping up&12.5\% (52.4\%)&37.5\% (36.9\%)\\
        Jumping forward&0.3\% (12.6\%)&25.3\% (25.7\%)\\
        Bowing&12.3\% (11.8\%)&62.7\% (25.1\%)\\
        Lying down&0.3\% (0.7\%)&12.5\% (25.0\%)\\
        Sitting down&25.2\% (36.2\%)&25.8\% (51.3\%)\\
        Turning around&25.4\% (35.7\%)&0.6\% (50.5\%)\\
        Forward rolling&25.4\% (48.3\%)&0.8\% (50.9\%)\\
        \bottomrule
    \end{tabular}
    \label{claude37-win}
\end{table}

\begin{table}[t]
    \centering
    \renewcommand\arraystretch{0.95}
    \caption{Pairwise agreement rate between crowd-user, experts, and MLLM judge (GPT-4o) on pairwise comparisons. The top and bottom are OpenAI's o3 versus Qwen3-32B and Gemini 2.0 Flash, respectively.}
   \begin{tabular}{lccc}
    \toprule                 
        Qwen3-32B&Expert 1&Expert 2&GPT-4o\\
        \midrule
        Crowd&84.3\%&86.2\%&82.7\%\\
        Expert 1&\textendash&94.9\%&85.8\%\\
        Expert 2&\textendash&\textendash&82.3\%\\
        \toprule                 
        Gemini 2.0 Flash&Expert 1&Expert 2&GPT-4o\\
        \midrule
        Crowd&88.7\%&87.2\%&86.7\%\\
        Expert 1&\textendash&96.2\%&86.3\%\\
        Expert 2&\textendash&\textendash&87.1\%\\
        \bottomrule
    \end{tabular}
    \label{agreement}
\end{table}

\begin{table}[t]
    \centering
    \renewcommand\arraystretch{0.95}
    \caption{O3's win-rate across crowd-user, experts, and GPT-4o judge on pairwise battles against Qwen3-32B and Gemini 2.0 Flash. `E' is the abbreviation for expert.}
   \begin{tabular}{lcccc}
    \toprule                 
        Competitor&Crowd&E1&E2&GPT-4o\\
        \midrule
        Qwen3-32B&61.8\%&63.2\%&64.7\%&61.3\%\\
        Gemini 2.0 Flash&66.9\%&68.8\%&68.4\%&65.5\%\\
        \bottomrule
    \end{tabular}
    \label{O3-win}
\end{table}

\noindent
{\bf Effects of Point-Light Quantity.} 
In Fig. \ref{few_points}, we visualize the walking motion of a man generated from the {\it top-3} models in BioMotion Arena using different numbers of points. 
It can be observed that none of them can perfectly render rational biological motion across a varying number of points. 
Among them, Gemini 2.5 Pro produces a relatively well-formed humanoid limb across various point configurations (except for the 8-point case, where the head was missing), while exhibiting the highest degree of motion smoothness. 
DeepSeek-R1 and o3 both produce varying degrees of joint misalignment and unnatural point clustering, resulting in irrational motion postures. Moreover, the results show that motion generation tasks involving either too many or too few point-lights pose a greater challenge to the models, suggesting the feasibility of such methods for more rigorous model evaluation.

\begin{figure}[t]
\begin{center}
\includegraphics[width=1\linewidth]{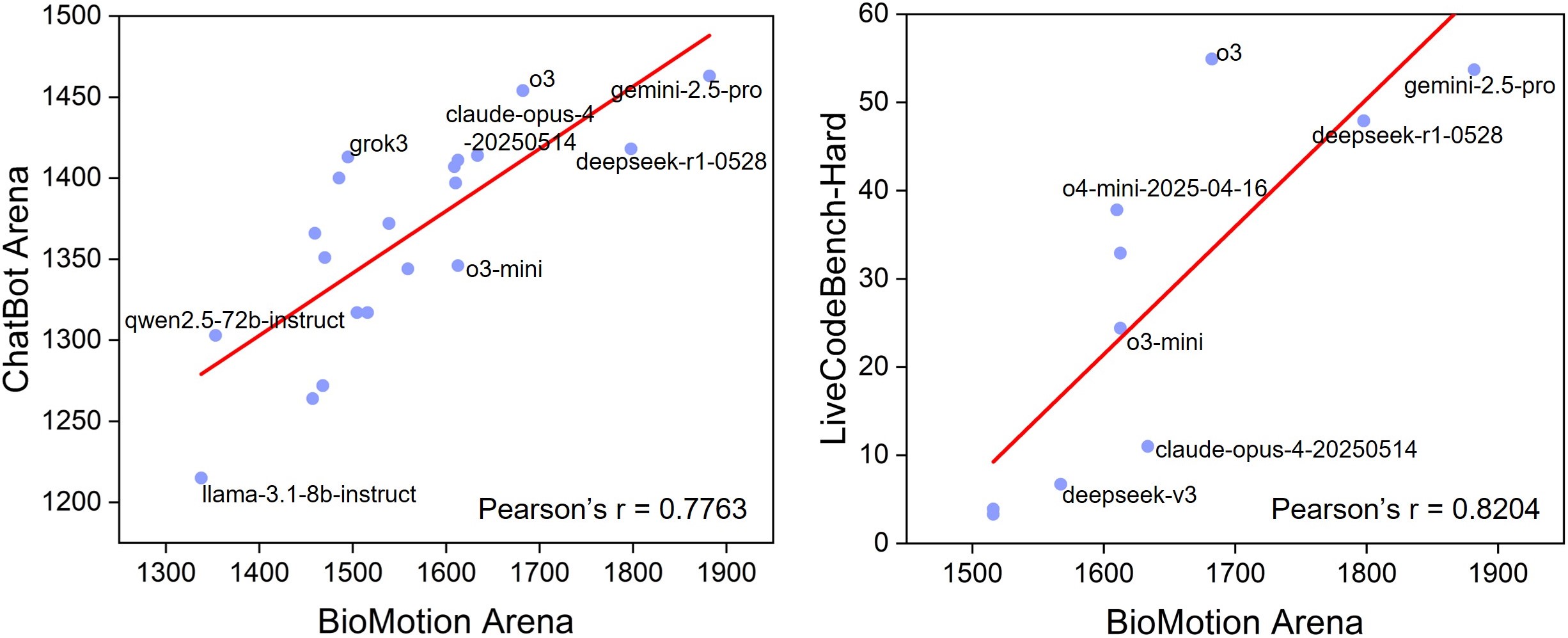}
\caption{The performance of models on different benchmarks, along with a best-fit line.}
\label{vschat-livecode}
\end{center}
\end{figure}

\subsection{Validating Vote Quality}
To evaluate the quality of crowdsourced votes, we randomly sampled 150 battles between OpenAI's o3 and Qwen3-32B, as well as o3 and Gemini 2.0 Flash, for scrutiny.
We invited two experts with research backgrounds in LLMs and asked them to annotate their preferences in each round of comparison. 
We adopted the same rating strategy as in crowdsourcing, where experts were given the target motions and votes blindly.
For reference, we also employed GPT-4o as a third-party judge. The agreement rates between crowd-users, experts, and the GPT-4o judge are shown in Tab. \ref{agreement}. The corresponding win-rates are presented in Tab. \ref{O3-win}.

To summarize, we observe high agreement rates (84.3\% to 88.7\%) between BioMotion Arena public participants and experts in both battles. Surprisingly, the agreement rates between two experts are extremely high and close (94.9\% and 96.2\%). 
As for the rest 5\% disagreements, we speculate that this was caused by either the generation of exceedingly poor point-light motion by both models or errors in the execution of the generated script.
Moreover, we find that GPT-4o has a uniformly high degree of consistency with crowd-users and experts, indicating the effectiveness of BioMotion Arena in conveying performance differences. The win-rates listed in Tab. \ref{O3-win} further demonstrate the quality of the preference data in BioMotion Arena.

\subsection{Comparison to Other Benchmarks.}
In Fig. \ref{vschat-livecode}, we compare BioMotion Arena against two widely used LLM benchmarks, ChatBot Arena \cite{chiang2024chatbot} and LiveCodeBench-Hard \cite{jainlivecodebench}, along with best-fit lines. 
We also compute the Pearson's correlation coefficient of model scores across the benchmarks, where BioMotion Arena owns a 0.7763 and 0.8204 correlation with ChatBot Arena and LiveCodeBench-Hard, respectively.
Based on the plots and correlations, we notice that although the overall performance trend of the two benchmarks aligns with that of the BioMotion Arena, some models have starkly contrasting performance. 
Specifically, claude-opus-4-20250514 performs substantially better on ChatBot Arena and relatively worse on LiveCodeBench-Hard compared to BioMotion Arena. 
In contrast, o3 achieves the top-1 performance in LiveCodeBench-Hard while exhibiting a little inferiority in BioMotion Arena.
We hypothesize that this discrepancy can be attributed to their differing evaluation mechanisms, i.e., LiveCodeBench is ground-truth-based, whereas the ChatBot Arena and our BioMotion Arena rely on open-ended subjective human preferences.

\section{Conclusion}
In this work, we present BioMotion Arena, a biologically inspired visual preference evaluation framework for large models, which enables intuitive, efficient, and easily perceptible performance comparison.
We conduct comparative insights across over 50 models by collecting over 45k human votes on 90 biological motion variants and utilize the Elo score to transfer the pairwise comparisons to relative scores, allowing for continuous updating of model performance.
Experimental results show that even cutting-edge reasoning LLMs or MLLMs fail to generate smooth and rational biological motions with a poor understanding of their kinetic-geometric patterns.
Additionally, we find a high agreement on preference votes from crowds and experts, demonstrating the superiority of BioMotion Arena in reflecting performance differences.
We plan to make this platform online to better collect human votes longitudinally and release more pairwise preference votes for future research.

\appendix
\newpage

\onecolumn

\begin{figure}
\centering
\includegraphics[width=0.9\linewidth]{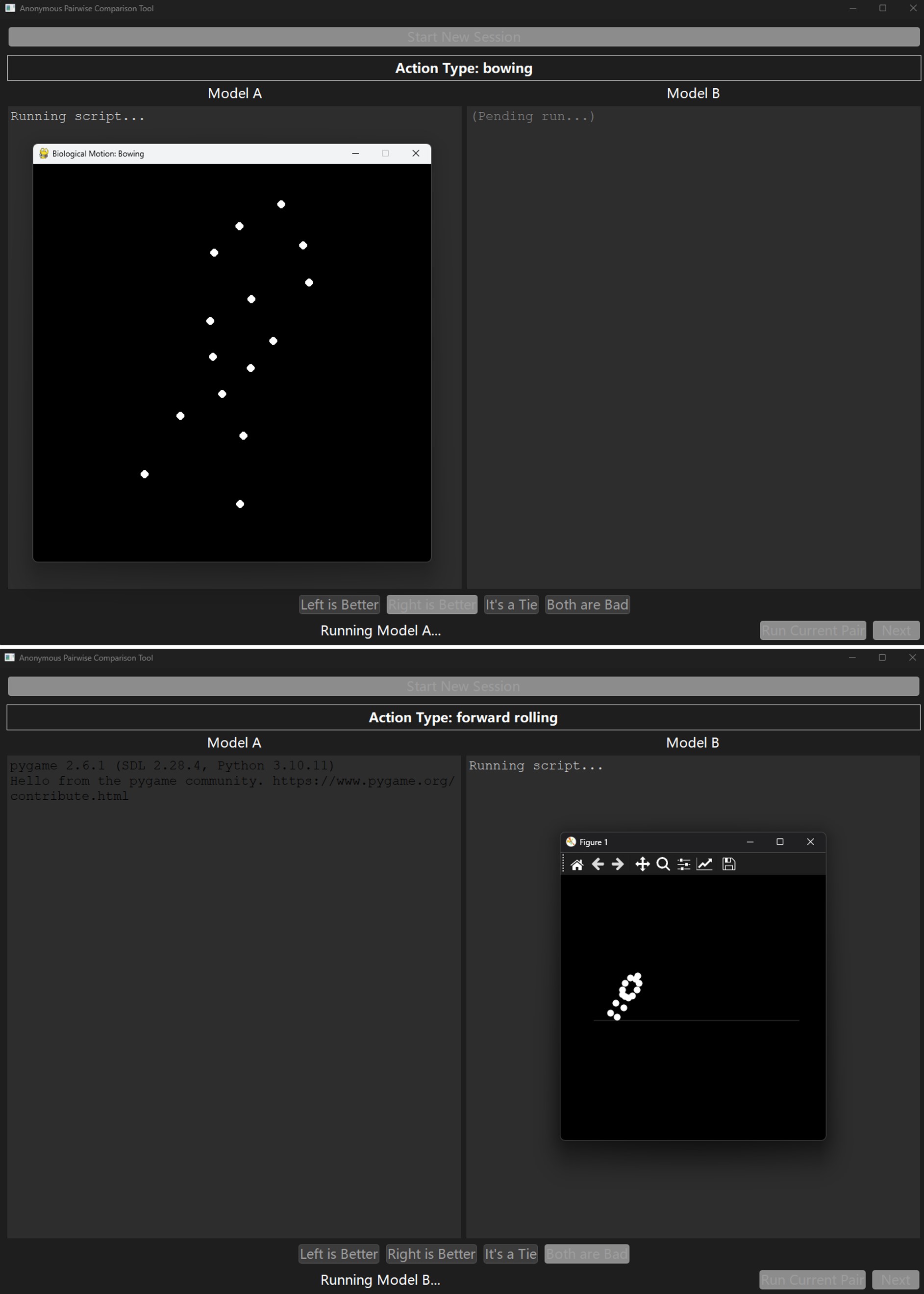}
\caption{Screenshot of the user interface for preference data collection.}
\label{UI}
\end{figure}

\begin{table*}[!h]
    \centering
    \renewcommand\arraystretch{0.8}
    \caption{List of models evaluated and their respective citations or websites. The upper and lower parts represent general large models and code-specific LLMs, respectively. `Thinking' and `Image' indicate that we evaluate with reasoning mode and also with image inputting, respectively.}
    \resizebox{1\linewidth}{!}{\begin{tabular}{llcc}
    \toprule                 
        {\bf Model Name}&{\bf Citation\textbackslash URL}&{\bf Thinking}&{\bf Image}\\
        \midrule
        \texttt{grok-4}&\url{https://docs.x.ai/docs/models/grok-4-0709}&\Checkmark&\Checkmark\\
        \texttt{grok-3-reasoning}&\url{https://x.ai/news/grok-3}&\Checkmark&\XSolidBrush\\
        \texttt{grok-3}&\url{https://x.ai/news/grok-3}&\XSolidBrush&\Checkmark\\
        \texttt{claude-opus-4-20250514-thinking}&\url{https://www.anthropic.com/news/claude-4}&\Checkmark&\Checkmark\\
        \texttt{claude-opus-4-20250514}&\url{https://www.anthropic.com/news/claude-4}&\XSolidBrush&\Checkmark\\
        \texttt{claude-3-7-sonnet-20250219}&\url{https://www.anthropic.com/news/claude-3-7-sonnet}&\XSolidBrush&\Checkmark\\
        \texttt{claude-3-5-sonnet-20241022}&\url{https://www.anthropic.com/news/3-5-models-and-computer-use}&\XSolidBrush&\Checkmark\\
        \texttt{gemini-2.5-pro}&\url{https://deepmind.google/models/gemini/pro/}&\Checkmark&\Checkmark\\
        \texttt{gemini-2.5-flash}&\url{https://deepmind.google/models/gemini/flash/}&\Checkmark&\Checkmark\\
        \texttt{gemini-2.0-flash}&\url{https://blog.google/technology/google-deepmind/google-gemini-ai-update-december-2024/}&\XSolidBrush&\Checkmark\\
        \texttt{gemini-1.5-pro}&\cite{team2024gemini}&\XSolidBrush&\Checkmark\\
        \texttt{gemini-1.5-flash}&\cite{team2024gemini}&\XSolidBrush&\Checkmark\\
        \texttt{deepseek-r1-250528}&\cite{guo2025deepseek}&\Checkmark&\XSolidBrush\\
        \texttt{deepseek-r1-250120}&\cite{guo2025deepseek}&\Checkmark&\XSolidBrush\\
        \texttt{deepseek-v3}&\cite{liu2024deepseek}&\XSolidBrush&\XSolidBrush\\
        \texttt{o4-mini-2025-04-16}&\url{https://openai.com/index/introducing-o3-and-o4-mini}&\Checkmark&\Checkmark\\
        \texttt{o3-2025-04-16}&\url{https://openai.com/index/introducing-o3-and-o4-mini}&\Checkmark&\Checkmark\\
        \texttt{o3-mini-2025-01-31}&\url{https://openai.com/index/openai-o3-mini}&\Checkmark&\XSolidBrush\\
        \texttt{o1-2024-12-17}&\url{https://openai.com/o1}&\Checkmark&\XSolidBrush\\
        \texttt{gpt-4.5-preview-2025-02-27}&\url{https://openai.com/index/introducing-gpt-4-5}&\XSolidBrush&\Checkmark\\
        \texttt{gpt-4.1-2025-04-14}&\url{https://openai.com/index/gpt-4-1}&\XSolidBrush&\Checkmark\\
        \texttt{gpt-4o-2024-11-20}&\cite{hurst2024gpt}&\XSolidBrush&\Checkmark\\
        \texttt{gpt-4o-mini-2024-07-18}&\cite{hurst2024gpt}&\XSolidBrush&\Checkmark\\
        \texttt{qwen3-32b}&\cite{yang2025qwen3}&\Checkmark&\XSolidBrush\\
        \texttt{qwen3-8b}&\cite{yang2025qwen3}&\Checkmark&\XSolidBrush\\
        \texttt{qwen2.5-vl-72b-instruct}&\cite{bai2025qwen25}&\XSolidBrush&\Checkmark\\
         \texttt{qwen2.5-vl-32b-instruct}&\cite{bai2025qwen25}&\XSolidBrush&\Checkmark\\
         \texttt{qwen2.5-vl-7b-instruct}&\cite{bai2025qwen25}&\XSolidBrush&\Checkmark\\
         \texttt{qwen2.5-vl-3b-instruct}&\cite{bai2025qwen25}&\XSolidBrush&\Checkmark\\
        \texttt{qwen2.5-72b-instruct}&\cite{qwen2.5}&\XSolidBrush&\XSolidBrush\\
        \texttt{qwen2.5-32b-instruct}&\cite{qwen2.5}&\XSolidBrush&\XSolidBrush\\
        \texttt{qwen2.5-7b-instruct}&\cite{qwen2.5}&\XSolidBrush&\XSolidBrush\\
        \texttt{qwen2-72b-instruct}&\cite{qwen2}&\XSolidBrush&\XSolidBrush\\
        \texttt{qwen2-7b-instruct}&\cite{qwen2}&\XSolidBrush&\XSolidBrush\\
        \texttt{gemma-3-4b-it}&\cite{team2025gemma}&\XSolidBrush&\Checkmark\\
        \texttt{gemma-2-2b-it}&\cite{team2024gemma}&\XSolidBrush&\XSolidBrush\\
        \texttt{llama-3.3-70b-instruct}&\url{https://www.llama.com/docs/model-cards-and-prompt-formats/llama3_3}&\XSolidBrush&\XSolidBrush\\
        \texttt{llama-3.2-3b-instruct}&\url{https://www.llama.com/docs/model-cards-and-prompt-formats/llama3_2}&\XSolidBrush&\XSolidBrush\\
        \texttt{llama-3.2-1b-instruct}&\url{https://www.llama.com/docs/model-cards-and-prompt-formats/llama3_2}&\XSolidBrush&\XSolidBrush\\
        \texttt{llama-3.1-70b-instruct}&\cite{grattafiori2024llama}&\XSolidBrush&\XSolidBrush\\
        \texttt{llama-3.1-8b-instruct}&\cite{grattafiori2024llama}&\XSolidBrush&\XSolidBrush\\
        \texttt{glm-4-9b-0414}&\cite{glm2024chatglm}&\XSolidBrush&\XSolidBrush\\
         \texttt{internvl3-78b-instruct}&\cite{zhu2025internvl3}&\XSolidBrush&\Checkmark\\
         \texttt{internvl3-8b-instruct}&\cite{zhu2025internvl3}&\XSolidBrush&\Checkmark\\
         \texttt{internvl2.5-78b-mpo}&\cite{chen2024expanding}&\XSolidBrush&\Checkmark\\
         \texttt{internvl2.5-38b-mpo}&\cite{chen2024expanding}&\XSolidBrush&\Checkmark\\
         \texttt{internvl2.5-8b-mpo}&\cite{chen2024expanding}&\XSolidBrush&\Checkmark\\
         \midrule
         \texttt{qwen2.5-coder-32b-instruct}&\cite{hui2024qwen2}&\XSolidBrush&\XSolidBrush\\
         \texttt{qwen2.5-coder-14b-instruct}&\cite{hui2024qwen2}&\XSolidBrush&\XSolidBrush\\
         \texttt{qwen2.5-coder-7b-instruct}&\cite{hui2024qwen2}&\XSolidBrush&\XSolidBrush\\
         \texttt{deepseek-coder-v2-lite-instruct}&\cite{zhu2024deepseek}&\XSolidBrush&\XSolidBrush\\
         \texttt{deepseek-coder-33b-instruct}&\cite{guo2024deepseek}&\XSolidBrush&\XSolidBrush\\
         \texttt{deepseek-coder-6.7b-instruct}&\cite{guo2024deepseek}&\XSolidBrush&\XSolidBrush\\
        \bottomrule
    \end{tabular}}
    \label{full_list}
\end{table*}

\newpage

\section{Limitations and Future Directions}
Although our BioMotion Arena provides a comprehensive evaluation platform for LLMs and MLLMs in visual biological motion generation tasks, there are several potential limitations acknowledged below.
First, since the model base in our BioMotion Arena is extensive, the number of battles for each model pair may incline to those popular one in future online collection, which results from the rapid updates and the potential user base of large models. This inclination may result in a biased distribution of scores. Therefore, optimizing the ranking algorithm is necessary to keep the arena fair.
Second, the current motion space may be limited, where only 10 typical human actions are included. This source might not accurately reflect the performance of large models in real-world environments or specialized domains, potentially leading to a skewed preference distribution.

In the future, we plan to make BioMotion Arena online, where public user can freely upload their specific biological motion generation requests, and develop comprehensive visual content leaderboards for both LLMs and MLLMs in less constrained ways, catering to more complex motions.

\section{Additional Details about BioMotion Arena}

A full list of models evaluated is provided in Tab. \ref{full_list}.
We demonstrate the user interface for preference data collection in Fig. \ref{UI}.
The complete action space and fine-grained attributes include:
\begin{itemize}
    \item Action: Walking, running, waving a hand, jumping up, jumping forward, bowing, lying down, sitting down, turning around, and forward rolling.
    \item Gender: Man, woman
    \item Happiness: Happy, sad
    \item Weight: Heavy, light 
\end{itemize}

\newpage
\section{Sample Prompts}
\label{subsec:clusters-example}
Below we provide the prompt format (with appropriate variants adding special tokens accommodating different motions used for code generation) in BioMotion Arena.

\noindent
\textbf{Example 1:} An example prompt of a basic biological motion ``A man is walking" for LLMs. We provide the generated results of Llama3.1-8B and OpenAI's o1, where the length of the code generated by the reasoning model o1 is twice that of Llama3.1-8B. We observe that o1 independently defined the movement patterns for 15 key points, while Llama 3.1-8B simply defined circular motion.

\noindent
\textbf{Example 2:} An example prompt of a fine-grained biological motion ``A happy man with heavy weight is walking" for LLMs, where we add the control of gender, mood, and weight. We provide the generated results of Qwen2.5-72B and DeepSeek-R1-20250528. 
We find that DeepSeek-R1-20250528 adopts the `pygame' to exhibit the target motion rather than matplotlib used in Qwen2.5-72B, resulting in more smooth motions that better meets the requirements.

\noindent
\textbf{Example 3:} An example prompt of a basic biological motion ``A man is walking" for MLLMs, where we include the reference image for input. We provide the generated results of InternVL2.5-38B and Gemini 2.5 Pro. We notice that the reasoning model, Gemini 2.5 Pro performed detailed torso, leg, and hips calculation with rigorous ankles movement functions, rather than simple point coordinates change in InternVL2.5-38B.

All the examples are listed in sequence as follows.

\newpage
\begin{tcolorbox}
\textbf{Example 1} \\
\scriptsize
\texttt{\textbf{<|System Prompt|>}} \\
\texttt{You are an expert Python programmer. You will be given a question (problem specification) and will generate a correct Python program that matches the specification. You will not return anything except for the program.} \\

\texttt{\textbf{<|User Prompt|>}} \\
\texttt{Question: write a Python program that shows a point-light stimulus animation which represents biological motion. Detailed Requirements: \\
1. Subject and Action: The animation depict a man is <action>. \\
2. Visual Style: The stimulus should consist of exactly 15 white point-lights moving against a solid black background.\\
3. Motion Quality: The animation must be realistic, coherent, and biomechanically plausible to accurately represent the specified human action. The movement should be smooth and natural.\\
}

\texttt{\textbf{<|Llama3.1-8B's Answer|>}} 
\begin{lstlisting}
```python
import numpy as np
import matplotlib.pyplot as plt
import matplotlib.animation as animation
import matplotlib.patches as patches

# Define the figure and axis
fig, ax = plt.subplots()

# Define the background color
ax.set_facecolor('black')

# Define the point-light properties
num_lights = 15
light_radius = 2
light_color = 'white'

# Define the initial positions of the point-lights
lights = []
for i in range(num_lights):
    light_x = np.sin(i * np.pi / num_lights) * 50 + 100
    light_y = np.cos(i * np.pi / num_lights) * 50 + 100
    lights.append([light_x, light_y])

num_frames = 100
fps = 30
duration = num_frames / fps

def movement(frame):
    # Update the positions of the point-lights
    for i in range(num_lights):
        light_x = np.sin((i * np.pi / num_lights) + frame / 10) * 50 + 100
        light_y = np.cos((i * np.pi / num_lights) + frame / 10) * 50 + 100
        lights[i] = [light_x, light_y]
    
    ax.clear()
    ax.set_facecolor('black')
    
    for light in lights:
        ax.add_patch(patches.Circle(light, light_radius, facecolor=light_color, edgecolor='black'))
    
    ax.set_xlim(0, 200)
    ax.set_ylim(0, 200)
    
    ax.set_xticks([])
    ax.set_yticks([])
    ax.set_frame_on(False)

ani = animation.FuncAnimation(fig, movement, frames=num_frames, interval=1000 / fps, repeat=False)

# Show the animation
plt.show()
```
\end{lstlisting}
\end{tcolorbox}

\begin{tcolorbox}
\scriptsize

\texttt{\textbf{<|OpenAI o1's Answer|>}} 
\begin{lstlisting}
```python
import numpy as np
import matplotlib.pyplot as plt
from matplotlib.animation import FuncAnimation
from math import sin, cos, pi

# Number of frames in one walking cycle
FRAMES_PER_CYCLE = 50
# Total frames (two walk cycles in this example)
TOTAL_FRAMES = FRAMES_PER_CYCLE * 2
# Amplitude of arm and leg swing (in radians)
ARM_SWING = pi / 6
LEG_SWING = pi / 4

# Define segment lengths
TORSO_LENGTH = 0.4
ARM_UPPER = 0.3
ARM_LOWER = 0.3
LEG_UPPER = 0.4
LEG_LOWER = 0.4

# Vertical base positions
HIP_HEIGHT = 0.8
SHOULDER_HEIGHT = HIP_HEIGHT + TORSO_LENGTH
HEAD_HEIGHT = SHOULDER_HEIGHT + 0.3

# Horizontal offsets
HIP_OFFSET = 0.1
SHOULDER_OFFSET = 0.2

def get_points(frame):
    """Return (x,y) coordinates for 15 point-lights representing a walking human."""
    # Time parameter for the walking cycle (0 to 2*pi)
    t = 2 * pi * (frame % FRAMES_PER_CYCLE) / FRAMES_PER_CYCLE

    # Angles for right arm/leg (and left out of phase by pi)
    rightArmAngle = ARM_SWING * sin(t)
    leftArmAngle = ARM_SWING * sin(t + pi)
    rightLegAngle = LEG_SWING * sin(t)
    leftLegAngle = LEG_SWING * sin(t + pi)

    # Define elbow/knee bending as smaller angles for a more lifelike motion
    rightElbowAngle = 0.5 * rightArmAngle
    leftElbowAngle = 0.5 * leftArmAngle
    rightKneeAngle = 0.5 * rightLegAngle
    leftKneeAngle = 0.5 * leftLegAngle

    points = []

    # 1) Head
    points.append((0.0, HEAD_HEIGHT))

    # 2) Neck
    neck = (0.0, SHOULDER_HEIGHT + 0.1 * (HEAD_HEIGHT - SHOULDER_HEIGHT))
    points.append(neck)

    # 3) Right Shoulder
    rShoulder = (SHOULDER_OFFSET, SHOULDER_HEIGHT)
    points.append(rShoulder)

    # 4) Right Elbow
    rElbowX = rShoulder[0] + ARM_UPPER * sin(rightArmAngle)
    rElbowY = rShoulder[1] - ARM_UPPER * cos(rightArmAngle)
    points.append((rElbowX, rElbowY))
\end{lstlisting}
\end{tcolorbox}

\begin{tcolorbox}
\scriptsize
\texttt{\textbf{<|OpenAI o1's Answer: Continuing from the previous paragraph...|>}} 
\begin{lstlisting}
```python
    # 5) Right Wrist
    rWristX = rElbowX + ARM_LOWER * sin(rightArmAngle + rightElbowAngle)
    rWristY = rElbowY - ARM_LOWER * cos(rightArmAngle + rightElbowAngle)
    points.append((rWristX, rWristY))

    # 6) Left Shoulder
    lShoulder = (-SHOULDER_OFFSET, SHOULDER_HEIGHT)
    points.append(lShoulder)

    # 7) Left Elbow
    lElbowX = lShoulder[0] + ARM_UPPER * sin(leftArmAngle)
    lElbowY = lShoulder[1] - ARM_UPPER * cos(leftArmAngle)
    points.append((lElbowX, lElbowY))

    # 8) Left Wrist
    lWristX = lElbowX + ARM_LOWER * sin(leftArmAngle + leftElbowAngle)
    lWristY = lElbowY - ARM_LOWER * cos(leftArmAngle + leftElbowAngle)
    points.append((lWristX, lWristY))

    # 9) Mid-Hip (center)
    midHip = (0.0, HIP_HEIGHT)
    points.append(midHip)

    # 10) Right Hip
    rHip = (HIP_OFFSET, HIP_HEIGHT)
    points.append(rHip)

    # 11) Right Knee
    rKneeX = rHip[0] + LEG_UPPER * sin(rightLegAngle)
    rKneeY = rHip[1] - LEG_UPPER * cos(rightLegAngle)
    points.append((rKneeX, rKneeY))

    # 12) Right Ankle
    rAnkleX = rKneeX + LEG_LOWER * sin(rightLegAngle + rightKneeAngle)
    rAnkleY = rKneeY - LEG_LOWER * cos(rightLegAngle + rightKneeAngle)
    points.append((rAnkleX, rAnkleY))

    # 13) Left Hip
    lHip = (-HIP_OFFSET, HIP_HEIGHT)
    points.append(lHip)

    # 14) Left Knee
    lKneeX = lHip[0] + LEG_UPPER * sin(leftLegAngle)
    lKneeY = lHip[1] - LEG_UPPER * cos(leftLegAngle)
    points.append((lKneeX, lKneeY))

    # 15) Left Ankle
    lAnkleX = lKneeX + LEG_LOWER * sin(leftLegAngle + leftKneeAngle)
    lAnkleY = lKneeY - LEG_LOWER * cos(leftLegAngle + leftKneeAngle)
    points.append((lAnkleX, lAnkleY))

    return np.array(points)

fig, ax = plt.subplots()
ax.set_facecolor('black')
ax.set_xlim(-1, 1)
ax.set_ylim(-0.2, 2)
ax.set_aspect('equal')
ax.axis('off')
initial_positions = get_points(0)
scatter = ax.scatter(initial_positions[:,0], initial_positions[:,1], c='white', s=20)

def update(frame):
    coords = get_points(frame)
    scatter.set_offsets(coords)
    return (scatter,)

ani = FuncAnimation(fig, update, frames=TOTAL_FRAMES, interval=50, blit=True)
plt.show()
\end{lstlisting}
\end{tcolorbox}

\newpage
\begin{tcolorbox}
\textbf{Example 2} \\
\scriptsize
\texttt{\textbf{<|System Prompt|>}} \\
\texttt{--- Same with the Example 1 ---} \\

\texttt{\textbf{<|User Prompt|>}} \\
\texttt{Question: write a Python program that shows a point-light stimulus animation which represents biological motion. Detailed Requirements: \\
1. Subject and Action: The animation depict a <happiness> + <gender> + with + <weight> + weight is <action>. \\
2. Visual Style: The stimulus should consist of exactly 15 white point-lights moving against a solid black background.\\
3. Motion Quality: The animation must be realistic, coherent, and biomechanically plausible to accurately represent the specified human action. The movement should be smooth and natural.\\
}

\texttt{\textbf{<|Qwen2.5-72B's Answer|>}} 
\begin{lstlisting}
```python
import numpy as np
import matplotlib.pyplot as plt
import matplotlib.animation as animation

# Define the positions of the 15 points for the walking man with a heavy weight
# These positions are simplified and should be adjusted for more realistic motion
points = [
    [0, 1], [0, 2], [0, 3],  # Head
    [-1, 0], [1, 0],         # Shoulders
    [-2, -1], [2, -1],       # Hips
    [-3, -2], [3, -2],       # Knees
    [-4, -3], [4, -3],       # Ankles
    [-1, -1], [1, -1],       # Elbows
    [-2, -2], [2, -2]        # Hands
]

# Convert to numpy array for easier manipulation
points = np.array(points)

# Function to update the position of the points for each frame
def update(frame):
    global points
    # Simulate walking by shifting the points horizontally
    shift = np.sin(frame / 10) * 0.5  # Sinusoidal motion for natural walking
    points[:, 0] += shift
    scatter.set_offsets(points)
    return scatter,

# Set up the plot
fig, ax = plt.subplots()
ax.set_xlim(-5, 5)
ax.set_ylim(-5, 5)
ax.set_facecolor('black')
scatter = ax.scatter(points[:, 0], points[:, 1], color='white')

# Create the animation
ani = animation.FuncAnimation(fig, update, frames=np.arange(0, 100), interval=50, blit=True)

# Display the animation
plt.show()
```
\end{lstlisting}
\end{tcolorbox}

\newpage
\begin{tcolorbox}
\scriptsize

\texttt{\textbf{<|DeepSeek-R1-20250528's Answer|>}} 
\begin{lstlisting}
```python
import pygame
import math
import sys
import numpy as np

# Initialize pygame
pygame.init()

# Constants
WIDTH, HEIGHT = 800, 600
BG_COLOR = (0, 0, 0)  # Black background
POINT_COLOR = (255, 255, 255)  # White points
POINT_RADIUS = 6
NUM_POINTS = 15
FPS = 60

# Joint indices (matching biological motion literature)
HEAD = 0
NECK = 1
SHOULDER_R = 2
SHOULDER_L = 3
ELBOW_R = 4
ELBOW_L = 5
WRIST_R = 6
WRIST_L = 7
HIP_R = 8
HIP_L = 9
KNEE_R = 10
KNEE_L = 11
ANKLE_R = 12
ANKLE_L = 13
FOOT_R = 14
FOOT_L = 15

# Set up display
screen = pygame.display.set_mode((WIDTH, HEIGHT))
pygame.display.set_caption("Biological Motion: Heavily Laden Walker")
clock = pygame.time.Clock()

class Walker:
    def __init__(self):
        self.scale = 100
        self.offset = np.array([WIDTH // 2, HEIGHT // 3])
        self.gravity = 0.6
        self.speed = 0.05
        self.t = 0
        self.foot_contact = [True, False]  # Right foot, left foot
        
    def get_joint_positions(self, phase):
        """Calculate realistic joint positions for heavy walker"""
        phase = phase % (2 * math.pi)
        
        # Core body motion (exaggerated for weight)
        torso_sway = 0.1 * math.sin(phase)
        vertical_bob = 0.5 * math.cos(2 * phase)  # Heavier bob
         # Body proportions
        positions = np.zeros((16, 2))  # More points than needed, we'll use 15
        
        # Head and torso
        positions[HEAD] = [0, -1.7 + vertical_bob]
        positions[NECK] = [0, -1.4 + vertical_bob]
        
        # Shoulders (wider for heavy person)
        positions[SHOULDER_R] = [0.7 - torso_sway, -1.3 + vertical_bob]
        positions[SHOULDER_L] = [-0.7 - torso_sway, -1.3 + vertical_bob]

\end{lstlisting}
\end{tcolorbox}

\begin{tcolorbox}
\scriptsize
\texttt{\textbf{<|DeepSeek-R1-20250528's Answer: Continuing from the previous paragraph...|>}} 
\begin{lstlisting}
```python
# Arms (shorter swing for heavy weight)
        arm_swing_r = 0.15 * math.sin(phase + math.pi/2)
        arm_swing_l = 0.15 * math.sin(phase + 3*math.pi/2)
        
        positions[ELBOW_R] = [0.7 - 0.4 - torso_sway, -0.8 + arm_swing_r]
        positions[ELBOW_L] = [-0.7 + 0.4 - torso_sway, -0.8 + arm_swing_l]
        
        # Hips (wider stance for heavy weight)
        positions[HIP_R] = [0.5 + torso_sway/2, 0 + vertical_bob]
        positions[HIP_L] = [-0.5 + torso_sway/2, 0 + vertical_bob]
        
        # Leg motion physics (shorter strides)
        leg_phase = phase if phase < math.pi else 2*math.pi - phase
        knee_angle = 1.2 - 0.4 * math.sin(leg_phase)
        thigh_r = 1.0 * math.sin(leg_phase - 0.2)
        thigh_l = 1.0 * math.sin(leg_phase + math.pi/2 + 0.2)

        positions[KNEE_R] = [
            0.5 + 0.3 * math.sin(phase) + torso_sway/2, 
            0.7 - thigh_r
        ]
        positions[ANKLE_R] = [
            positions[KNEE_R][0] + 0.2 * math.sin(phase/2),
            positions[KNEE_R][1] + 0.8 + thigh_r/2
        ]
        positions[KNEE_L] = [
            -0.5 + 0.3 * math.sin(phase + math.pi) + torso_sway/2, 
            0.7 - thigh_l
        ]
        positions[ANKLE_L] = [
            positions[KNEE_L][0] + 0.2 * math.sin((phase+math.pi)/2),
            positions[KNEE_L][1] + 0.8 + thigh_l/2
        ]
        
        # Feet placement
        positions[FOOT_R] = [positions[ANKLE_R][0] + 0.2, positions[ANKLE_R][1] + 0.1]
        positions[FOOT_L] = [positions[ANKLE_L][0] + 0.2, positions[ANKLE_L][1] + 0.1]
        
        return positions * self.scale + self.offset

    def update(self):
        """Update walker's state"""
        self.t += self.speed
        return self.get_joint_positions(self.t)[:NUM_POINTS]

# Create walker
walker = Walker()

# Main loop
running = True
while running:
    for event in pygame.event.get():
        if event.type == pygame.QUIT:
            running = False

    # Clear screen
    screen.fill(BG_COLOR)
    
    # Get and draw points
    points = walker.update()
    for point in points:
        pygame.draw.circle(screen, POINT_COLOR, (int(point[0]), int(point[1])), POINT_RADIUS)
    
    # Update display
    pygame.display.flip()
    clock.tick(FPS)

pygame.quit()
sys.exit()
\end{lstlisting}
\end{tcolorbox}

\newpage
\begin{tcolorbox}
\textbf{Example 3} \\
\scriptsize
\texttt{\textbf{<|System Prompt|>}} \\
\texttt{--- Same with the Example 1 ---} \\

\texttt{\textbf{<|User Prompt|>}} \\
\texttt{Question: Question: Given an example image <image>, write a Python program that shows a point-light stimulus animation which represents biological motion. Detailed Requirements: \\
1. Subject and Action: The animation depict a man is <action>. \\
2. Visual Style and 3. Motion Quality: --- Same with the Example 1 ---\\
}

\texttt{\textbf{<|InternVL2.5-38B's Answer|>}} 
\begin{lstlisting}
```python
import matplotlib.pyplot as plt
import matplotlib.animation as animation
import numpy as np
class WalkerAnimation:
    def __init__(self):
        self.fig, self.ax = plt.subplots()
        self.lights = plt.scatter([], [])
        self.ax.set_xlim(0, 10)
        self.ax.set_ylim(0, 10)
        self.ax.set_aspect('equal')
        self.ax.set_facecolor('black')
        self.ax.axis('off')
        self.trajectories = self.generate_trajectories()
    def generate_trajectories(self):    # Define coordinates for 15 points
        points = np.zeros((15, 2))  # Initial position of the points
        points[0] = (5.0, 5.0)     # Head
        points[1] = (4.5, 4.0)     # Left shoulder
        points[2] = (5.5, 4.0)     # Right shoulder
        points[3] = (4.0, 3.0)     # Left elbow
        points[4] = (6.0, 3.0)     # Right elbow
        points[5] = (3.5, 2.0)     # Left hand
        points[6] = (6.5, 2.0)     # Right hand
        points[7] = (5.0, 2.0)     # Left hip
        points[8] = (5.0, 2.0)     # Right hip
        points[9] = (4.5, 1.0)     # Left knee
        points[10] = (5.5, 1.0)    # Right knee
        points[11] = (4.0, 0.0)     # Left foot
        points[12] = (6.0, 0.0)     # Right foot
        points[13] = (4.8, 3.0)     # Left torso point
        points[14] = (5.2, 3.0)     # Right torso point
        trajectories = []
        for i in range(100):
            frame = points.copy()
            # Simulate walking motion
            alpha = 0.1 * np.sin(i * 0.1)
            beta = 0.1 * np.cos(i * 0.1)
            frame[7][0] += 0.02
            frame[8][0] += 0.02
            frame[9][0] += 0.03 + alpha
            frame[10][0] += 0.03 - alpha
            frame[11][0] += 0.05 + beta
            frame[12][0] += 0.05 - beta
            frame[9][1] += -0.03
            frame[10][1] += -0.03
            frame[11][1] += -0.05
            frame[12][1] += -0.05
            trajectories.append(frame)
        return trajectories
    def animate(self, i):
        self.lights.set_offsets(self.trajectories[i])
        return self.lights,
    def show(self):
        ani = animation.FuncAnimation(self.fig, self.animate, frames=len(self.trajectories), interval=50, blit=True)
        plt.show()
if __name__ == "__main__":
    animation = WalkerAnimation()
    animation.show()
```
\end{lstlisting}
\end{tcolorbox}

\newpage
\begin{tcolorbox}
\scriptsize
\texttt{\textbf{<|Gemini 2.5 Pro's Answer|>}} 
\begin{lstlisting}
```python
import pygame
import math

def main():
    """
    Main function to set up and run the Pygame animation.
    """
    # --- Initialization ---
    pygame.init()

    # --- Screen and Display Settings ---
    screen_width = 800
    screen_height = 600
    screen = pygame.display.set_mode((screen_width, screen_height))
    pygame.display.set_caption("Biological Motion: Walking")

    # --- Colors and Appearance ---
    BLACK = (0, 0, 0)
    WHITE = (255, 255, 255)
    DOT_RADIUS = 5

    # --- Animation Control ---
    clock = pygame.time.Clock()
    fps = 60
    time_angle = 0.0  # Represents the current phase of the walking cycle
    walk_speed = 0.05 # Rate at which the time_angle increments

    # --- Biomechanical Model Parameters ---
    # These parameters control the amplitude and characteristics of the motion
    # for different body parts. They are tuned to create a natural-looking walk.
    
    # Center of the screen for positioning the figure
    center_x = screen_width / 2
    center_y = screen_height / 2
    
    # Torso motion
    sway_amplitude = 8          # Side-to-side sway of the torso
    bob_amplitude = 4           # Up-and-down bob of the torso
    
    # Body proportions
    pelvis_base_y = 30          # Base vertical position of the pelvis from center
    torso_length = 60           # Distance from pelvis to sternum
    head_offset_y = 70          # Distance from sternum to head
    shoulder_width = 35         # Half the distance between shoulders
    hip_width = 25              # Half the distance between hips
    hip_offset_y = 10           # Vertical distance from pelvis center to hip joints
    
    # Leg motion
    leg_swing_x = 40            # Horizontal amplitude of ankle swing
    foot_lift_y = 30            # Vertical amplitude of ankle lift during swing
    knee_bend_y = 25            # Extra vertical lift of the knee to simulate bending
    knee_clearance_x = 10       # Horizontal movement of the knee to clear the other leg
    floor_level_y = 200         # The "ground" plane relative to the center
    
    # Arm motion
    arm_swing_x = 25            # Horizontal amplitude of wrist swing
    arm_swing_y = 8             # Vertical amplitude of wrist swing
    elbow_bend_x = 18           # Outward bend of the elbow
    arm_base_y = 55             # Base vertical position of the wrist relative to shoulder

```
\end{lstlisting}
\end{tcolorbox}

\newpage
\begin{tcolorbox}
\scriptsize
\texttt{\textbf{<|Gemini 2.5 Pro's Answer: Continuing from the previous paragraph...|>}} 
\begin{lstlisting}
```python
def get_joint_positions(t):
        """
        Calculates the 2D screen positions of the 15 joints for a given time angle t.
        The model creates a frontal view of a person walking on a treadmill.
        
        Args:
            t (float): The current phase angle of the walking cycle (0 to 2*pi).
            
        Returns:
            list: A list of (x, y) tuples for the 15 joint positions.
        """
        points = {}

        # --- Torso Calculation ---
        # The torso sways side-to-side (sin(t)) and bobs up and down (cos(2t)).
        sway = sway_amplitude * math.sin(t)
        bob = bob_amplitude * math.cos(2 * t)
        
        points['pelvis'] = (center_x + sway, center_y + pelvis_base_y + bob)
        points['sternum'] = (points['pelvis'][0], points['pelvis'][1] - torso_length)
        points['head'] = (center_x + sway * 0.5, points['sternum'][1] - head_offset_y)

        # --- Leg Calculation ---
        # Right leg's motion is based on t, left leg is offset by pi (half a cycle).
        phase_r_leg = t
        phase_l_leg = t + math.pi
        
        # Hips are positioned relative to the pelvis.
        points['r_hip'] = (points['pelvis'][0] - hip_width, points['pelvis'][1] + hip_offset_y)
        points['l_hip'] = (points['pelvis'][0] + hip_width, points['pelvis'][1] + hip_offset_y)
        
        # Ankles swing back and forth. The foot lifts off the ground during the swing phase.
        # max(0, sin(phase)) creates a curve that is positive during the first half
        # of the cycle (swing) and zero during the second half (stance).
        r_ankle_lift = foot_lift_y * max(0, math.sin(phase_r_leg))
        l_ankle_lift = foot_lift_y * max(0, math.sin(phase_l_leg))
        
        points['r_ankle'] = (points['r_hip'][0] + leg_swing_x * math.cos(phase_r_leg), center_y + floor_level_y - r_ankle_lift)
        points['l_ankle'] = (points['l_hip'][0] + leg_swing_x * math.cos(phase_l_leg), center_y + floor_level_y - l_ankle_lift)
        
        # Knees are positioned between the hip and ankle, with an added bend.
        # This bend pulls the knee up and slightly out to clear the other leg.
        r_knee_bend = knee_bend_y * max(0, math.sin(phase_r_leg))
        l_knee_bend = knee_bend_y * max(0, math.sin(phase_l_leg))
        r_knee_clearance = knee_clearance_x * max(0, math.sin(phase_r_leg))
        l_knee_clearance = knee_clearance_x * max(0, math.sin(phase_l_leg))

        points['r_knee'] = (
            (points['r_hip'][0] + points['r_ankle'][0]) / 2 - r_knee_clearance,
            (points['r_hip'][1] + points['r_ankle'][1]) / 2 - r_knee_bend
        )
        points['l_knee'] = (
            (points['l_hip'][0] + points['l_ankle'][0]) / 2 + l_knee_clearance,
            (points['l_hip'][1] + points['l_ankle'][1]) / 2 - l_knee_bend
        )

        phase_r_arm = t + math.pi
        phase_l_arm = t

        points['r_shoulder'] = (points['sternum'][0] - shoulder_width, points['sternum'][1])
        points['l_shoulder'] = (points['sternum'][0] + shoulder_width, points['sternum'][1])

```
\end{lstlisting}
\end{tcolorbox}

\newpage
\begin{tcolorbox}
\scriptsize
\texttt{\textbf{<|Gemini 2.5 Pro's Answer: Continuing from the previous paragraph...|>}} 
\begin{lstlisting}
```python
        points['r_wrist'] = (
            points['r_shoulder'][0] + arm_swing_x * math.sin(phase_r_arm),
            points['r_shoulder'][1] + arm_base_y + arm_swing_y * math.cos(phase_r_arm)
        )
        points['l_wrist'] = (
            points['l_shoulder'][0] + arm_swing_x * math.sin(phase_l_arm),
            points['l_shoulder'][1] + arm_base_y + arm_swing_y * math.cos(phase_l_arm)
        )
        
        # Elbows are positioned between the shoulder and wrist, with an outward bend.
        # The bend is greatest when the arm is at the side (cos(phase) is 0).
        r_elbow_bend = elbow_bend_x * (1 - abs(math.cos(phase_r_arm)))
        l_elbow_bend = elbow_bend_x * (1 - abs(math.cos(phase_l_arm)))
        
        points['r_elbow'] = (
            (points['r_shoulder'][0] + points['r_wrist'][0]) / 2 - r_elbow_bend,
            (points['r_shoulder'][1] + points['r_wrist'][1]) / 2
        )
        points['l_elbow'] = (
            (points['l_shoulder'][0] + points['l_wrist'][0]) / 2 + l_elbow_bend,
            (points['l_shoulder'][1] + points['l_wrist'][1]) / 2
        )
        
        # The 15 points in order are: Head, Sternum, Pelvis, L/R Shoulder, L/R Elbow,
        # L/R Wrist, L/R Hip, L/R Knee, L/R Ankle.
        # The following list ensures a consistent drawing order.
        ordered_points = [
            points['head'], points['sternum'], points['pelvis'],
            points['r_shoulder'], points['l_shoulder'],
            points['r_elbow'], points['l_elbow'],
            points['r_wrist'], points['l_wrist'],
            points['r_hip'], points['l_hip'],
            points['r_knee'], points['l_knee'],
            points['r_ankle'], points['l_ankle']
        ]
        return ordered_points

    # --- Main Loop ---
    running = True
    while running:
        # Event handling
        for event in pygame.event.get():
            if event.type == pygame.QUIT or (event.type == pygame.KEYDOWN and event.key == pygame.K_ESCAPE):
                running = False

        # Update state
        time_angle = (time_angle + walk_speed) % (2 * math.pi)

        # Get current joint positions
        joint_positions = get_joint_positions(time_angle)

        # Drawing
        screen.fill(BLACK)
        for pos in joint_positions:
            pygame.draw.circle(screen, WHITE, (int(pos[0]), int(pos[1])), DOT_RADIUS)

        # Refresh screen
        pygame.display.flip()

        # Control frame rate
        clock.tick(fps)

    pygame.quit()

if __name__ == '__main__':
    main()
```
\end{lstlisting}
\end{tcolorbox}

\end{document}